%% file: paper.tex
\RequirePackage{etoolbox}
\newtoggle{arxiv}
\toggletrue{arxiv}
\newtoggle{final}
\iftoggle{arxiv}{%
    \documentclass[a4paper,11pt]{article}
    \toggletrue{final}
    \usepackage{arxiv}
    \usepackage[numbers,sort]{natbib}
    \newcommand{\cvspace}[1]{}
}{%
    \documentclass[11pt]{article}
    \usepackage[review]{acl}
    \newcommand{\cvspace}[1]{\vspace{#1}}
}
\usepackage{times}
\usepackage{latexsym}
\usepackage[T1]{fontenc}
\usepackage[utf8]{inputenc}
\usepackage{microtype}
\usepackage{inconsolata}
\usepackage{graphicx}
\usepackage{soul}
\PassOptionsToPackage{table}{xcolor}

\input{common}

\newcommand{\abox}[1]{\iftoggle{arxiv}{#1}{\mbox{#1}}}

\title{%
    \abox{Mixture of Weight-shared Heterogeneous
    \mbox{Group Attention Experts}}
    for \mbox{Dynamic Token-wise KV Optimization}}
\renewcommand{\shorttitle}{}

\author{%
    Guanghui Song \\
    \small University of Chinese Academy of Sciences; \\
    \small Shenzhen Institutes of Advanced Technology, \\
    \small Chinese Academy of Sciences \\
    \texttt{gh.song@siat.ac.cn} \\ \And
    Dongping Liao \\
    \small University of Macau \\
    \texttt{yb97428@um.edu.mo} \\ \And
    Yiren Zhao \\
    \small Imperial College London \\
    \texttt{a.zhao@imperial.ac.uk} \\ \And
    Kejiang Ye \\
    \small Shenzhen Institutes of Advanced Technology, \\
    \small Chinese Academy of Sciences; \\
    \small Shenzhen University of Advanced Technology \\
    \texttt{kj.ye@siat.ac.cn} \\ \And
    Cheng-zhong Xu \\
    \small University of Macau \\
    \texttt{czxu@um.edu.mo} \\ \And
    Xitong Gao\thanks{Corresponding author: \texttt{xt.gao@siat.ac.cn}} \\
    \small Shenzhen Institutes of Advanced Technology, \\
    \small Chinese Academy of Sciences; \\
    \small Shenzhen University of Advanced Technology \\
    \texttt{xt.gao@siat.ac.cn}
}

\begin{document}

\maketitle

\input{abstract}

\input{intro}
\input{related}
\input{method}
\input{results}
\input{conclusion}
\iftoggle{arxiv}{%
\bibliographystyle{acl_natbib}
}{%
\input{limitations}
}
\bibliography{references}
\clearpage
\appendix
\input{appendix}

\end{document}

%% file: common.tex
\usepackage{hyperref}
\usepackage{url}
\usepackage{booktabs}
\usepackage{amsfonts}
\usepackage{nicefrac}
\usepackage{microtype}
\usepackage{mathtools,bm,amssymb}
\usepackage{xcolor}
\usepackage{hyperref}
\usepackage{subcaption,placeins}
\usepackage{adjustbox,multirow,dcolumn}
\usepackage{algorithm,algpseudocode,setspace}
\usepackage{xstring}
\usepackage{cleveref}
\usepackage{graphicx}
\usepackage{tikz}
\usepackage{pgfplots}
\usepackage{comment}
\usepackage{framed}

\makeatletter

\input{math_commands}

\DeclareRobustCommand\onedot{\futurelet\@let@token\@onedot}
\def\@onedot{\ifx\@let@token.\else.\null\fi}

\newcommand{\ie}{\emph{i.e\@\onedot}}

\newcommand{\first}{1\textsuperscript{st}}
\newcommand{\second}{2\textsuperscript{nd}}
\newcommand{\third}{3\textsuperscript{rd}}
\newcommand{\ordinal}[1]{{#1}\textsuperscript{th}}

\graphicspath{{figures/}}

\usetikzlibrary{external}
\usepgfplotslibrary{colorbrewer}
\pgfplotsset{compat=newest}
\newcolumntype{d}{D{.}{.}{3.2}}
\newcolumntype{B}{>{\boldmath\DC@{.}{.}{3.2}}c<{\DC@end}}

\newcommand{\MethodVerb}{{mixSGA}}
\newcommand{\Method}{\emph{\MethodVerb}}

\DeclarePairedDelimiter{\parens}{\lparen}{\rparen}

\DeclarePairedDelimiter{\bracks}{[}{]}
\DeclarePairedDelimiter{\braces}{\{}{\}}

\DeclarePairedDelimiter{\ceils}{\lceil}{\rceil}

\newcommand{\x}{{\mathbf{x}}}
\newcommand{\z}{{\mathbf{z}}}

\newcommand{\ratios}{\bm{\rho}}
\newcommand{\ratio}{\rho}
\newcommand{\mask}{\mathbf{m}}

\newcommand{\keyweight}{{\mathbf{w}^\textrm{k}}}
\newcommand{\valueweight}{{\mathbf{w}^\textrm{v}}}
\newcommand{\keybias}{{\mathbf{b}^\textrm{k}}}
\newcommand{\valuebias}{{\mathbf{b}^\textrm{v}}}
\newcommand{\routeweight}{\bm\phi}
\newcommand{\routebias}{\bm\beta}

\newcommand{\realset}{\mathbb{R}}

\DeclareMathOperator{\topk}{\mathsf{top}}

\DeclareMathOperator{\loss}{\mathcal{L}}
\DeclareMathOperator{\sceloss}{\mathcal{L}^\textrm{sce}}
\DeclareMathOperator{\auxloss}{\mathcal{L}_\textrm{aux}}
\DeclareMathOperator{\modelloss}{\mathcal{L}_\textrm{model}}

\DeclareMathOperator{\indicator}{\mathbf{1}}

\DeclareMathOperator{\sigmoid}{{\sigma}}
\DeclareMathOperator{\softmax}{\mathsf{softmax}}
\DeclareMathOperator{\routescore}{\mathsf{S}}

\newcommand{\upar}{\ensuremath{\uparrow}}
\newcommand{\downar}{\ensuremath{\downarrow}}
\renewcommand{\r}[3]{{#1}\textrm{:}{#2}\textrm{:}{#3}}

\newcommand{\routeprefill}{\mathbf{T}_\textrm{prefill}}
\newcommand{\routedecode}{\mathbf{T}_\textrm{decode}}

\makeatother

%% file: math_commands.tex
\usepackage{amsmath,amsfonts,bm}

\def\1{\bm{1}}

\DeclareMathAlphabet{\mathsfit}{\encodingdefault}{\sfdefault}{m}{sl}
\SetMathAlphabet{\mathsfit}{bold}{\encodingdefault}{\sfdefault}{bx}{n}

\DeclareMathOperator*{\argmax}{arg\,max}

%% file: abstract.tex
\begin{abstract}
    Transformer models face scalability challenges
    in causal language modeling (CLM)
    due to inefficient memory allocation
    for growing key-value (KV) caches,
    which strains compute and storage resources.
    Existing methods like Grouped Query Attention (GQA)
    and token-level KV optimization improve efficiency
    but rely on rigid resource allocation,
    often discarding ``low-priority'' tokens or statically grouping them,
    failing to address the dynamic spectrum of token importance.
    We propose \Method{},
    a novel mixture-of-expert (MoE) approach
    that dynamically optimizes token-wise computation and memory allocation.
    Unlike prior approaches,
    \Method{} retains all tokens while adaptively routing them
    to specialized experts with varying KV group sizes,
    balancing granularity and efficiency.
    Our key novelties include:
    \textbf{(1)} a token-wise expert-choice routing mechanism
    guided by learned importance scores,
    enabling proportional resource allocation
    without token discard;
    \textbf{(2)} weight-sharing across grouped attention projections
    to minimize parameter overhead;
    and \textbf{(3)} an auxiliary loss
    to ensure one-hot routing decisions
    for training-inference consistency
    in CLMs.
    Extensive evaluations
    across Llama3, TinyLlama, OPT, and Gemma2 model families
    show \Method{}'s superiority over static baselines.
    On instruction-following and continued pretraining tasks,
    \Method{} achieves higher ROUGE-L and lower perplexity
    under the same KV budgets.
\end{abstract}

%% file: intro.tex
\section{Introduction}\label{sec:intro}

Transformer architectures
have emerged as the backbone of modern deep learning,
powering state-of-the-art advancements across diverse fields
such as natural language processing \cite{vaswani2023attentionneed},
computer vision \cite{alexey2020image},
reinforcement learning \cite{parisottoefficient}
and beyond \cite{le2020dual, chen2023personalized}.
However,
their self-attention mechanism,
while effective,
suffers from quadratic computational and memory costs
with respect to the sequence length,
posing scalability challenges \cite{10.1145/3530811}.
\input{figures/motivation.tex}\unskip

Efforts towards addressing these challenges
have largely focused on improving
the efficiency of attention mechanisms.
One line of methods
seek to improve the design of the attention block.
Grouped Query Attention (GQA) \cite{ainslie2023gqa},
for example,
reduces computational overhead
by clustering keys and values
into coarse groups,
which reduces the number of processed KV pairs.
Nevertheless,
GQA assumes static group sizes
and allocates resources uniformly,
disregarding variations in token importance.
Some works
have been devoted to optimize the memory footprint
of the widely adopted KV cache \cite{waddington2013kv}.
Token-level approaches,
such as DynamicKV \cite{zhou2024dynamickv},
introduce flexible KV cache allocation
by prioritizing high-value tokens.
However,
these methods often involve rigid resource allocation strategies
that neglect to fully exploit the significance of low-priority tokens.

Another promising line of work
\cite{shazeer2017,lepikhin2020gshard}
adopts MoEs to dynamically route tokens
to a subset of experts,
enabling efficient resource utilization.
While these approaches
achieve computational efficiency,
they frequently suffer from imbalanced expert utilization.
Moreover,
their coarse-grained routing
overlooks token-level variability,
highlighting the need for finer-grained adaptivity
in token-level resource allocation.
In many cases,
tokens deemed less important
are outright discarded
or receive minimal processing,
which can lead to degraded performance
for certain tasks.

Our work is motivated by the experimental findings
presented in \Cref{fig:motivation},
which reveal that token importance exhibits dynamic behavior
and spans a wide spectrum.
This observation
naturally inspires the high-level idea
of tailoring experts' token selection based on their importance.
While this approach holds significant promise
for efficiently leveraging the potential of token prioritization,
it faces several critical challenges that must be addressed.
First,
current token-choice routing (TCR) approaches
can result in unbalanced expert utilization,
particularly challenging to tune for heterogeneous capacities of experts,
as tokens may always prefer high capacity experts,
posing the risk of collapsed routing mechanisms.
Second,
existing expert-choice routing (ECR) methods
shifts imbalanced expert utilization
to token utilization,
where tokens may be assigned to multiple experts,
while some tokens are ignored.
Third,
existing ECRs also introduce \emph{training and inference disparities}
in CLMs,
where during the training or prefill phase,
routings are made based on the complete sequence,
whereas the decoding-phase routings
are made based on past context.


To surmount these obstacles,
we introduce \Method{}.
Unlike prior work
that discards less significant tokens,
our method retains all tokens
while dynamically allocating computation and memory resources
proportionally to their importance.
For the experts,
we propose a weight-sharing mechanism
across grouped attentions,
allowing the model to remain lightweight
while dynamically scaling
based on token significance.
To overcome the routing disparity
between prefill and decode stages,
we propose a layer-wise auxiliary loss
that encourages routing consistency.


Our contributions are summarized as follows:
\begin{itemize}
    \item \textbf{The \Method{} Framework}:
    \Method{} integrates dynamic token-wise routing
    with KV attention head grouping,
    enabling adaptive computational/memory allocation
    without discarding tokens.
    It also uses weight-sharing for parameter efficiency.

    \item \textbf{Autoregressive Expert-Choice Routing}:
    We propose a novel past-context routing mechanism
    with an auxiliary loss to ensure prefill-decode consistency
    in CLMs.
    It also enables flexible tuning of individual expert capacities.

    \item \textbf{Broad Empirical Validation}:
    We demonstrates superior efficiency
    and performance over static and dynamic baselines
    across OPT, Llama3 and Gemma2 models
    on diverse instruction-following
    and continued pretraining benchmarks.
\end{itemize}


%% file: figures/motivation.tex
\begin{figure}[t]
    \centering
    \begin{subfigure}[b]{0.48\linewidth}
        \centering
        \includegraphics[
            width=\iftoggle{arxiv}{0.7}{1}\linewidth,
            trim=12pt 12pt 10pt 12pt, clip
        ]{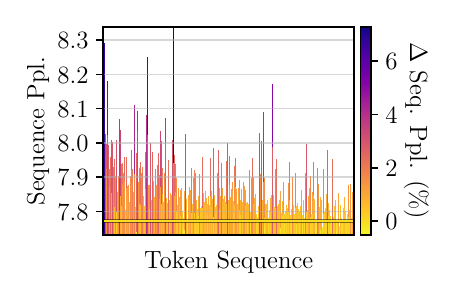}\cvspace{-5pt}
        \caption{%
            Change for each token.
        }\label{fig:motivation:sequence}
    \end{subfigure}\hfill%
    \begin{subfigure}[b]{0.48\linewidth}
        \centering
        \includegraphics[
            width=\iftoggle{arxiv}{0.7}{1}\linewidth,
            trim=12pt 12pt 12pt 12pt, clip
        ]{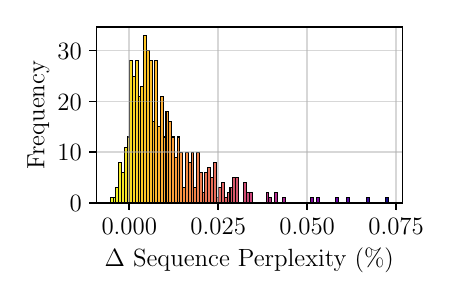}\cvspace{-5pt}
        \caption{%
            Change distribution.
        }\label{fig:motivation:histogram}
    \end{subfigure}\cvspace{-0.5em}
    \caption{%
        \textbf{Token importance
        is dynamic and has a wide spectrum.}
        We replace each token's forward pass
        with multi-query attention
        on Llama3.1-8b
        by averaging key and value states across heads,
        and see the sequence perplexity changes
        on a sample sequence of~WikiText-2.
    }\label{fig:motivation}
    \cvspace{-1em}
\end{figure}

%% file: related.tex
\section{Related Work}\label{sec:related}

\subsection{KV Cache Management}

KV cache optimization
enhances the memory efficiency of CLMs \cite{waddington2013kv}.
Recent methods
such as PyramidKV \cite{cai2024pyramidkv},
DynamicKV \cite{zhou2024dynamickv},
H2O \cite{zhang2024h2o}
and NACL \cite{nacl2024}
aim to reduce memory footprint,
typically by prioritizing high-utility tokens
based on heuristics, random, or learned importance,
and evict those deemed less critical
to stay within a constrained memory budget.
Furthermore,
methods like SnapKV \cite{li2024snapkv}
and FastGen \cite{ge2024discard}
focus on pattern- and importance-based token selection
to optimize KV cache efficiency
during inference.

Despite these improvements,
such methods often rely on predefined grouping mechanisms,
which may not fully capture token-level variability.
As a result,
they can overlook the nuanced importance of individual tokens,
limiting their ability
to optimize resource utilization effectively.
In addition,
 inevitably remove tokens
from the attention context,
which may lead to degraded performance
on fine-grained contextual understanding.
By contrast,
\Method{} adaptively allocates smaller KV cache sizes
to less critical tokens without evicting them,
striking a balance between efficiency
and preserving full contextual integrity.
It also preserves all tokens,
and instead of hard eviction,
adaptively allocates memory and compute resources
in proportion to token importance.
This design ensures that even less critical tokens
retain their contextual influence,
offering a more flexible and context-preserving alternative
to hard cache eviction.

\subsection{MoE Routing Strategies and Challenges}

MoEs provide a scalable approach to increase model capacity
without proportionally increasing computational costs
\cite{shazeer2017,lepikhin2020gshard}.
In token-choice routing (TCR) MoE models
such as GShard \cite{lepikhin2020gshard}
and Switch Transformer \cite{fedus2022switch},
each token independently select an expert or a subset of experts
for resource-efficient computation.
As TCR allows each token to independently select an expert,
it may suffer from imbalanced expert utilization,
inefficient resource allocation,
and potentially unstable training dynamics.
This is especially problematic
when experts have heterogeneous capacities,
as tokens may favor experts with higher capacities,
making it challenging to balance expert loads.
Expert-choice routing (ECR) \cite{zhou2022moe}
enables experts to select tokens for processing
while explicitly defining their capacities,
improving load balancing and resource utilization.
Despite its advantages over TCR,
ECR presents two significant challenges
in the context of CLMs:
\textbf{(1)} it requires access to the entire input sequence
to make routing decisions,
which is incompatible with CLMs
that rely solely on past tokens to predict the next token;
and \textbf{(2)} it shifts the issue
of \emph{imbalanced} expert utilization
to \emph{imbalanced} token utilization,
where some tokens may remain unprocessed by any expert,
while others may be redundantly processed by multiple experts.

Our work differentiates itself
from existing TCR and ECR methods
by introducing a routing mechanism
specifically designed for CLMs.
This mechanism evaluates token significance
based on partial sequence context,
enables dynamic expert selection,
and ensures prefill-decode routing consistency
in decoder-only architectures.
Additionally,
our method accommodates experts with heterogeneous capacity,
delivering fine-grained resource allocation
and improved efficiency on computation and memory costs.

\subsection{Grouped Attention Methods}

Grouped Query Attention (GQA) \cite{ainslie2023gqa}
reduces the computational and memory costs
by merging keys and values into larger groups,
reducing the number of KV pairs
processed during attention.
This can lead to inefficiencies
when token importance varies significantly,
as structured merging
fail to prioritize tokens critical to the task.
\iftoggle{arxiv}{}{%
Decoupled-Head Attention (DHA) \cite{chen2024dha}
adaptively merges attention heads across layers,
while Align Attention \cite{alignattn2024}
uses \( \ell_0 \) regularization
to convert multi-head attention
into group-based formulations.
Mixture-of-Head Attention (MoH) \cite{moh2024}
reformulates attention heads as experts in a MoE,
employing TCR for sparse head activation.}
Cross-layer Attention (CLA) \cite{brandon2024cla}
merges key and value projectors across adjacent layers.
While these approaches
learn to enhance structural efficiency,
they rely on static group sizes
for attention heads during inference,
assuming uniform token importance
and lacking fine-grained, token-level adaptability.
Additionally,
these methods do not support heterogeneous expert configurations
with varying group sizes,
limiting their adaptability.

In contrast,
\Method{} integrates the strengths
of grouped attention and token-level adaptivity
by dynamically routing each token
to weight-shared experts
with heterogeneous KV configurations,
based on learned token importance.
Unlike prior methods,
\Method{} retains all tokens,
ensuring no loss of contextual information,
while adaptively allocating computational and memory resources
at both group and token levels.

%% file: method.tex
\section{The \Method{} Method}\label{sec:method}

\Method{},
\underline{mix}ture of weight-\underline{s}hared
\underline{g}rouped \underline{a}ttention experts,
combines dynamic token-wise expert assignment
with token-level KV optimization
to achieve efficient attention computation and minimize KV memory.
This section elaborates on the key components,
including the routing mechanism for expert selection,
the mixture of weight-shared KV grouping experts,
and the auxiliary loss designed
to improve prefill/decode consistency.
\input{figures/method_overview}

\subsection{Prefill and Training Phase Routing}\label{sec:method:prefill}

\paragraph{Token-to-expert mapping score function}
Given an input sequence \( X \in \realset^{L \times D} \),
where \( L \) is the sequence length
and \( D \) is the embedding dimension,
we define the following token-to-expert mapping scoring function
for all tokens,
a trainable linear layer
\( \routescore \colon \realset^{L \times D} \to \realset^{L \times E} \)
with weight \( \routeweight \in \realset^{D \times E} \)
and bias \( \routebias \in \realset^{E} \),
where \( E \) is the number of experts:
\begin{equation}
    \routescore(\x) = \sigmoid\parens{\x \routeweight + \routebias},
\end{equation}
and \( \sigmoid(\cdot) \) is the sigmoid function.
The sigmoid activation
ensures bounded scores within \( [0, 1] \),
avoiding additional normalization during training.

\paragraph{MoEs with Heterogeneous Capacities}
To facilitate downstream KV cache optimization,
our method employs a routing mechanism
that dynamically assigns tokens to experts
based on predefined capacity ratios.
These ratios regulate token distribution among experts,
aligning with memory and computational constraints.
Assume that we have \( E \) experts,
where with predefined capacity ratios for each expert
\( \ratios = \braces{\ratio_1, \ratio_2, \dots, \ratio_E} \),
representing the fraction of tokens it processes.
The capacity ratios lie in the range \( [0, 1] \),
and are normalized such that the sum of all ratios is 1,
\ie{}, \( {\small\sum}_{e=1}^E \ratio_e = 1 \).
During training,
our token-to-expert routing
thus takes the scoring function output \( \routescore(\x) \)
and greedily assigns tokens to experts progressively.
For the \ordinal{\( e \)} expert,
we assign tokens based on the top-\( \ceils{\ratio_e L} \) scores,
and route the remaining tokens
to the next (\ordinal{\(e + 1\)}) expert.
Formally,
it employs the following sparse masking function
\(
    \mask_e\colon
        \realset^{L \times E} \to \braces{0, 1}^{L \times E}
\),
where:
\begin{equation}\label{eq:mask}
    \mask_e\parens{\x} = \indicator\bracks*{
        \topk_{\ceils{\ratio_e L}} \parens*{
            \routescore(\x) {\textstyle\prod}_{i = 1}^{e - 1} (1 - \mask_i\parens{\x})
        }
    },
\end{equation}
and \( \indicator \) denotes the element-wise indicator function,
producing 1 for the top-\( \ceils{\ratio_e L} \) scores,
and 0 otherwise.
Note \( \mask_e(\x) \)
depends on the masks of preceding experts,
ensuring that tokens previously
assigned to other experts are skipped,
thereby guaranteeing an exclusive mapping
of each token to a single expert.


\subsection{Decode-Phase Routing}\label{sec:method:decode}

The preceding paragraphs
outline the training/prefill phase of our token-wise ECR mechanism,
which operates on a sequence of tokens as input.
However,
this routing approach
cannot be directly applied
to the decoding phase of CLMs,
where tokens are generated iteratively,
this means that we need a different routing strategy
for the decode phase.

A key advantage of \cref{eq:mask}
is that it ensures exclusive expert mapping for each token,
resulting in \( {\sum}_{e=1}^E \mask_e(\x) \)
being a one-hot vector for each token.
If we encourage both phases
to have the same expert assignments,
we can simply use \( \arg\max\routescore(\x) \)
to determine the expert assignment
during decoding.
During the decoding phase,
expert assignments for the next token
are then determined
by simply taking the \( \arg\max \) of the scoring function,
\ie{},
This approach eliminates the need for a top-\( k \) operation
over the entire input sequence,
which is infeasible during decoding.
To summarize,
the prefill and decode phases
use the following routing functions:
\begin{equation}
    \begin{aligned}
    \routeprefill(\x) = {} & {\textstyle\sum}_{e=1}^E \mask_e(\x), \\
    \routedecode(\x) = {} & \indicator\bracks*{\argmax\parens{\routescore(\x)} = e}.
    \end{aligned}
\end{equation}

\subsection{Prefill-Decode Consistency Loss}\label{sec:method:loss}

To align \( \arg\max\routescore(\x) \)
with the expert assignment
\( \arg\max\routeprefill(\x) \),
we introduce the following consistency loss
where \( \argmax\routeprefill(\x) \)
extracts the expert index assigned to each token:
\begin{equation}
    \auxloss(\x) = \sceloss\parens[\big]{
        \routescore(\x), \argmax\mathbf{T}(\x)
    }.
\end{equation}

The total training loss for the model
combines the primary language-modeling loss \( \mathcal{L}_{\text{model}} \)
with the auxiliary loss \( \mathcal{L}_{\text{aux}}(\x^{(l)}) \)
applied across all layers \( l \in \braces{1, \ldots, L} \),
weighted by \( \alpha \):
\begin{equation}\textstyle
    \loss = \modelloss + \frac{\alpha}{L} \sum_{l=1}^{L} \auxloss(\x^{(l)}).
\end{equation}

\subsection{%
    Mixture of Weight-Shared GQAs
}\label{sec:method:mogqa}

\paragraph{KV projection}
Building on the token-wise expert assignment
described earlier,
we extend the attention mechanism
by introducing a mixture of \emph{weight-shared} GQAs.
Each expert processes its assigned tokens independently
and maintains KV caches
tailored to its group configuration,
achieving an efficient trade-off
between computation and memory.
Assuming a pretrained attention layer with \(
    (\keyweight, \valueweight) \in \realset^{D \times D},
    (\keybias, \valuebias) \in \realset^{H \times D},
\) key and value weights and biases,
where 
\( D \) is the embedding dimension,
we first define the following key and value projection
\( p^j_h \colon \realset^{L \times D} \to \realset^{H \times L \times (D / H
)} \)
for the \ordinal{\( h \)} head,
where \( j \in \braces{\mathrm{k}, \mathrm{v}} \),
\( h \in \braces{1, \ldots, H} \),
and:
\begin{equation}\textstyle
    P^{\,j}(\x)_h = \parens*{
        \mathbf{w}^j \x^\top + \mathbf{b}^{\,j}
    }_{
        \bracks*{\small \frac{D(h - 1)}{H} + 1 \, : \, \frac{Dh}{H}}
    },
\end{equation}
Here,
the subscript \( \z_{\bracks{a:b}} \)
denotes the slice operation
which selects elements from the first dimension of \( \z \)
ranging from \( a \) to \( b \).

\paragraph{KV grouping}
Inspired by GQA \cite{ainslie2023gqa},
for each expert \( f^j_e \),
we design the following mechanism
to reduce the number of projected KV heads
from \( H \) to \( H / 2^e \) groups
of size \( 2^e \)
by taking the average of the corresponding grouped heads.
Specifically,
for each grouping \( g \in G_e \) of expert \( e \),
we have
\(
    f^j_e \colon
        \realset^{H \times L \times (D / H)}
        \to \realset^{{H / 2^e} \times L \times (D / H)}
\):
\begin{equation}
    f^j_{e,g} (\x) = \nicefrac{1}{2^{e}} \,
        {\textstyle\sum}_{h \in g} p^{\,j}(\x)_h,
\end{equation}
where \( G_e \) groups
a range of heads by size \( 2^e \),
For example,
if \( H = 4 \) and \( E = 3 \),
we have \( G_1 = \braces{\braces{1}, \braces{2}, \braces{3}, \braces{4}} \),
\( G_2 = \braces{\braces{1, 2}, \braces{3, 4}} \),
and \( G_3 = \braces{\braces{1, 2, 3, 4}} \).
Notably to ensure parameter efficiency,
we share the same key and value weights
across all experts.
While for mathematical clarity
we define the mean operation
over the projected heads,
one can easily instead aggregate the KV projection weights
before applying the projection operation
to achieve the same effect.

Due to this grouping,
the total KV cache size
is thus adjusted
based on which expert processes the token,
with the cache size of the \ordinal{\( e \)} expert
being \( H / 2^e \) of the original size.

\paragraph{Attention computation}
Before computing the attention,
for expert \( e \)
we match the KV head counts \( H / 2^e \)
with the query head count \( H \)
by repeating the KV heads \( 2^e \) times
using \(
    h^j_{e,g} \colon
        \realset^{H / 2^e \times L \times (D / H)}
        \to \realset^{H \times L \times (D / H)}
\):
\begin{equation}
    h^j_{e,g}(\x) = \textstyle
        f^j_{e,g}(\x) \otimes \mathbf{1}_{2^e}.
\end{equation}
where \( \otimes \) denotes the outer product,
and \( \mathbf{1}_{2^e} \) is a vector of ones of size \( 2^e \).
Finally,
the overall result computed by the MoE is:
\begin{equation}\textstyle
    h^j(\x) = \sum_{e=1}^E \mask_e(\x) \odot h^j_{e,g}(\x).
\end{equation}
It is noteworthy that since \( \mask_e(\x) \) is sparse
and has token-wise exclusive expert assignment,
the most of the \( h^j_{e,g}(\x) \) are zeroed out and skipped.
In practice,
this is carried out efficiently
with scatter and gather tensor operations.

The attention computation is then performed
following the standard scaled dot-product attention mechanism,
where \( q(\x) \) is the original query projection:
\begin{equation}
    a(\x) = \softmax\parens*{
        { q(\x) h^{\text{k}}(\x)^\top } / {\sqrt{D}}
    } h^{\text{v}}(\x).
\end{equation}

%

\paragraph{Expert Allocation for Memory Efficiency}
\Method{} computes varying KV sizes per token
thanks to its dynamic routing mechanism
assigning tokens to experts of different group sizes.
For \( E = 3 \) experts,
the group sizes are \( 1, 2, 4 \) respectively,
and the head counts are thus \( H, H / 2, H / 4 \).
This means that on average given a ratio of \( a:b:c \),
all tokens require
\( \parens{a + b / 2 + c / 4} / \parens{a + b + c} \)
of the original KV size.
Along with the KV cache,
we also store a single index value
for each token to track expert assignment.

\paragraph{Integration with KV eviction}
Although \Method{}
dynamically allocates per-token KV sizes,
it remains fully compatible with KV eviction
such as H2O \cite{zhang2024h2o} and NACL \cite{nacl2024}
to further reduce memory usage.

%% file: figures/method_overview.tex
\begin{figure}[t]
    \centering%
    \includegraphics[
        width=\iftoggle{arxiv}{0.7}{1}\linewidth, trim={0pt 170pt 410pt 0pt}
    ]{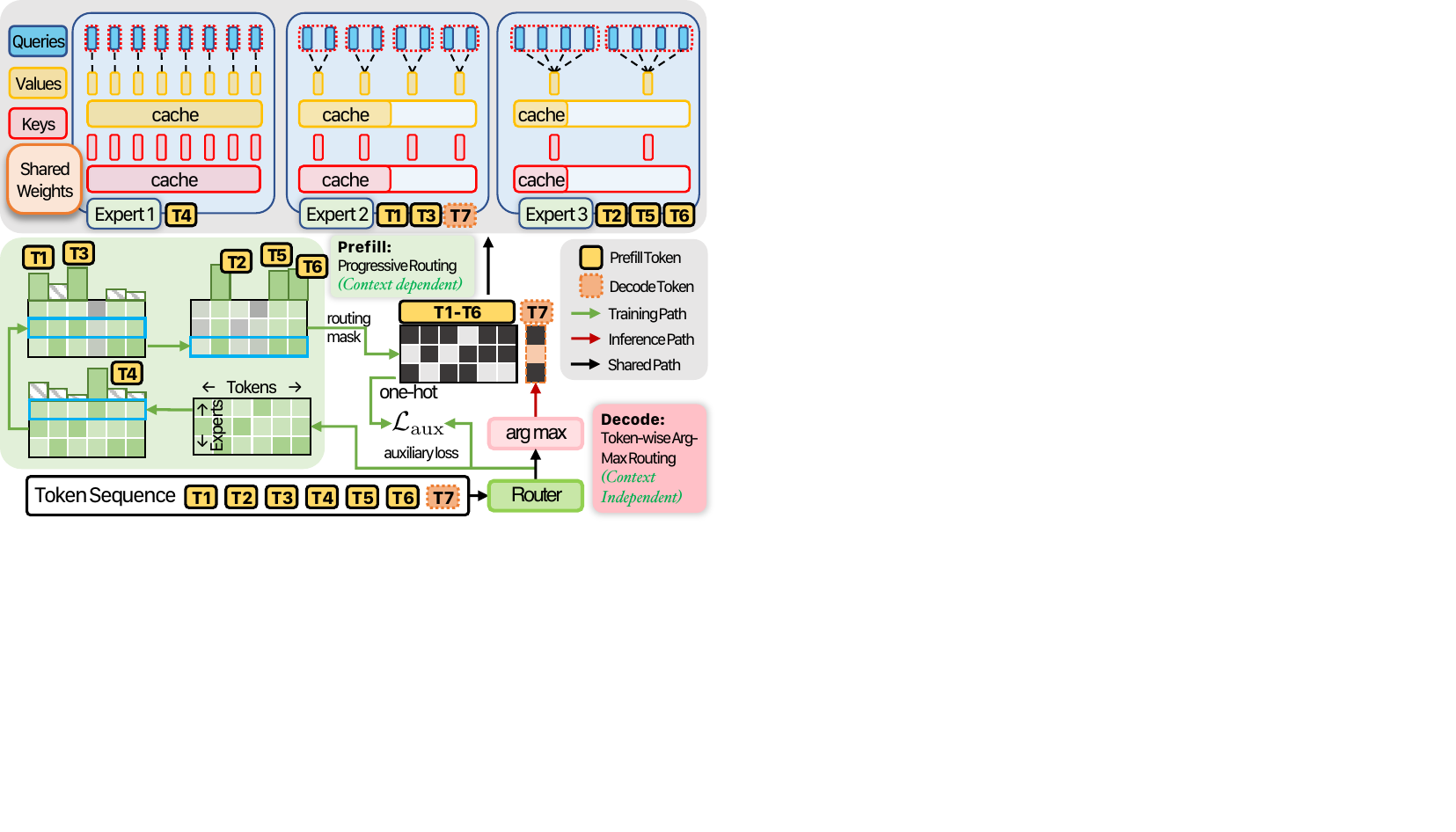}\cvspace{-0.5em}
    \caption{%
        High-level overview of \Method{}.
        During training,
        the router learns to compute assignment scores
        for each token-expert pair.
        These scores are utilized to sequentially route \( \ratio_e \) tokens
        to the \ordinal{\( e \)} expert,
        ordered by their computational and memory costs,
        while leaving the remaining tokens to be routed to other experts.
        The experts consist of a set of key and value projections
        that generate state representations at varying levels of granularity.
        This process ensures a unique routing assignment for each token,
        which is subsequently used to encourage the router
        to produce one-hot decisions through an auxiliary loss.
        During decoding,
        each token independently selects its corresponding expert
        through \( \arg\max \).
    }\label{fig:overview}\cvspace{-1em}
\end{figure}

%% file: results.tex
\section{Experiments}\label{sec:result}


\subsection{Supervised Fine-tuning}\label{sec:result:sft}

\paragraph{Models and methods}
We evaluate \Method{} on the following CLMs:
OPT-\{125m,355m\} \cite{zhang2022opt},
Llama3.1-8b, Llama3.2-\{1b,3b\} \cite{touvron2023llama},
and Gemma2-2b \cite{gemma2},
covering various model sizes and architectures.
As a default baseline,
we implement a GQA-variant of the original models
which forms KV head groups of size 2
by initializing the KV projection matrices
with the mean of the group.
For fair comparisons,
\Method{} is configured with expert density ratios
which maintain the same active KV head counts,
and thus the same KV size, as GQA\@.
It keeps the pretrained weights from the original models,
and randomly initializes the newly added routing weights
with He initialization \cite{he2015init}
and biases with zeros.

\paragraph{Training and evaluation setup}
We fine-tune the modified models
on the Dolly-15k instruction-following dataset
\cite{DatabricksBlog2023DollyV2}
with 14,000 training samples,
and evaluate their performance on 5 conversational datasets:
Dolly (DL, 500 testing samples from Dolly-15k),
Self-Instruct (SI) \cite{wang2023si},
Vicuna (VC) \cite{vicuna2023},
Super-Natural Instructions (SN) \cite{wang2022sn},
and Unnatural Instruction (UI) \cite{honovich2023un}.
In addition to the ROUGE-L (R-L) scores,
which measure the longest common sub-sequence
between generated and reference answers,
we also evaluate all answers to the queries
using DeepSeek-V3 \cite{deepseekai2024dsv3}
to provide feedback scores ranging from 0 to 10.
The template to generate feedback is provided in \Cref{app:setup}.
All hyperparameter configurations
are provided in \Cref{app:setup} for reproducibility.
\input{tables/main}

\paragraph{Main Results}

For supervised fine-tuning tasks,
we initiate our approach
by conducting a grid search on a smaller model (OPT-355M)
to determine the optimal expert density ratios,
incrementing by \( 0.1 \)
while maintaining the total KV size constant
at \( 50\% \) of the original model.
Our results show that allocating tokens
as 30\% to experts with a group size of 1,
10\% to size 2,
and 60\% to size 4
optimizes performance across most metrics.
This \r316 ratio consistently outperforms other configurations.
As shown in \Cref{tab:main},
\Method{} consistently outperforms GQA
across various benchmarks and model sizes.
These results demonstrate \Method{}'s ability
to dynamically allocate resources
and improve performance over static GQA baselines.

\subsection{Continued Pretraining}\label{sec:result:continued}

\paragraph{Models and methods}
We investigate \Method{}'s ability
in continued pretraining on additional corpus.
We used a TinyLlama-1.1B model \cite{zhang2024tinyllama},
which was pretrained on SlimPajama \cite{cerebras2023slimpajama}
and StarCoder \cite{li2023starcoder}
and adapted its weights
to GQA with group size set to 2,
CLA \cite{brandon2024cla},
and \Method{}.
Both CLA and \Method{}
aligns the same KV cache size as the GQA baseline.

\paragraph{Training and evaluation setup}
We train the models with each method applied
for one epoch of MiniPile \cite{kaddour2023minipile},
which amounts to 1.6 billion tokens.
We use a diverse set of benchmarks
to evaluate the resulting models:
HellaSwag \cite{zellers2019hellaswag},
PIQA \cite{Bisk2020},
Winogrande \cite{sakaguchi2019winogrande},
ARC-Easy (ARC-E),
ARC-Challenge (ARC-C) \cite{allenai_arc},
and the perplexity on Wikitext-2 \cite{merity2016pointer}.
For the first six tasks,
higher accuracy (\%) indicates better performance,
while lower perplexity on Wikitext-2
reflects stronger language modeling ability.
The training and evaluation details
are provided in \Cref{app:setup}.

\paragraph{Main Results}
In our continued pretraining setting,
the key challenge is to recover previously learned capabilities
of the model with a fraction of data
drawn from a distribution domain similar to the original pretraining data.
As shown in \Cref{tab:tiny_llama_results},
\Method{}
consistently demonstrates competitive or superior accuracy
on most benchmarks.
It attains 37.00\% on HellaSwag
and 56.30\% on Winogrande,
both surpassing GQA (group size = 2) and CLA.
Performance on ARC-C (25.17\%)
also exceeds that of the baselines,
highlighting \Method{}'s strength
in handling more challenging tasks.
\Method{} also shows a clear advantage in Wikitext-2 PPL,
delivering the lowest value (20.46) among all models.
To summarize,
these results indicate that \Method{}
can enable the model to preserve previously acquired knowledge,
as applying it to existing models
does not impact their pretrained weights.
\input{tables/tiny_llama_results}
\input{tables/mixsga_h2o}

\paragraph{\Method{} compliments cache eviction better}\label{sec:result:h2o}
To investigate the compatibility of \Method{}
with dynamic KV cache eviction strategies,
we conduct a set of controlled experiments
by integrating H2O~\cite{zhang2024h2o}
with both GQA and \Method{} on Gemma2-2b.
These experiments are designed to evaluate
whether the orthogonal benefits
of token-level eviction and token-wise KV allocation
can be combined effectively.
Both GQA and \Method{}
are configured to operate under a shared KV budget
of 50\% of the original size,
with H2O applied as a post-processing eviction method
to further compress memory.
We vary the H2O keep ratio from 80\% down to 20\%
to simulate increasing memory pressure.
The results,
shown in \Cref{tab:mixsga_h2o},
demonstrate that \Method{} consistently outperforms GQA
across all compression levels.
This validates that \Method{}
not only preserves the contextual coherence
lost in aggressive token eviction,
but also enhances the effectiveness of cache compression
when used in conjunction with existing methods like H2O.
The results demonstrate that integrating \Method{}
with cache eviction policies
further enhances its applicability in inference tasks
while reducing KV memory footprint.

\subsection{Ablation Studies}\label{sec:result:ablation}

To comprehensively attribute the impact
of each component in \Method{},
we perform ablation studies
under three key aspects by varying the following:
expert density ratios and expert counts,
and the auxiliary loss with learned routing mechanism.
Experiments in \Cref{%
    tab:ablation:ratios,%
    tab:ablation:expert_count}
and \Cref{tab:ablation:routing}
are conducted on Llama3.2-1b and Gemma2-2B respectively,
following the same setup in \Cref{sec:result:sft}.
We provide detailed analyses of the results below.

\paragraph{Varying the expert ratios}
\Cref{tab:ablation:ratios}
investigates the effect of varying density ratios among experts
while keeping a fixed KV size budget of \(50\%\).
We systematically increase the ratio
assigned to the \second{} expert in a group of size 2,
testing configurations from \r112 to \r192,
Our results reveals that evaluation metrics
improve as the \second{} expert's ratio decreases,
indicating a preference for allocating more tokens
to the \first{} and \third{} experts.
This suggests the model
prioritizes assigning important tokens to the \first{} expert,
which retains the original model's KV projection weights,
while routing less significant tokens
to the smallest (\third{}) expert.
\input{tables/effect_ratios}

\paragraph{Varying the expert counts}
In \Cref{tab:ablation:expert_count},
we investigate the influence
of employing 2-3 experts
while maintaining a fixed total KV budget of \( 50\% \).
Specifically,
we compare configurations
with \r316, \r340, \r112, \r120 ratios.
Here,
a value of \( 0 \) for the \third{} expert
indicates its exclusion from the model.
Remarkably,
we observe that introducing a \third{} expert
significantly enhances performance,
achieving an average ROUGE-L score improvement
of up to 3.12 across all benchmarks.
Given the variable information content of individual tokens,
this finding highlights the critical role of the \third{} expert
in capturing less crucial tokens within the input sequence,
allowing the other two experts
to focus on processing more significant ones.
\input{tables/effect_kv_cache}


\paragraph{Learned Routing}
To assess the auxiliary loss and learned routing mechanism,
we conduct experiments on Gemma2-2B
with a \r316 expert ratio,
following \Cref{sec:result:sft}.
As shown in \Cref{tab:ablation:routing},
we found that removing the auxiliary loss
leads to inconsistent routing
between prefill and decoding,
resulting in near-random expert assignments
(\r{0.3458}{0.3306}{0.3236} for the 3 experts on Dolly),
as the model never learns to route according to expert density ratios.
This causes a severe average ROUGE-L drop (21.20 to 7.35).
We also found that replacing the learned router
with a router that randomly assigns experts
per the \r316 ratio degrades performance.
\input{tables/auxiliary_loss}

\paragraph{Varying KV Budgets}
To evaluate the influence of varying KV budgets
on language modeling ability,
we conducted comparative experiments
involving \Method{}, GQA, and CLA
across different KV budgets
using the TinyLlama continued pretraining task
as outlined in \Cref{sec:result:continued}.
For \Method{},
the configurations were set as follows:
\r001 for a \( 25\% \) KV budget,
\r118 targeting \( 35\% \),
\r316 for \( 50\% \),
and \r110 for \( 75\% \).
CLA was configured to align with these KV sizes.
Given that the TinyLlama-1.1B attention module
comprises only 4 heads,
GQA could thus only employ a group size of 2
to achieve a \( 50\% \) KV budget.

\begin{figure}[t]
    \centering
    \includegraphics[
        width=\iftoggle{arxiv}{0.6}{1}\linewidth,
        trim=12pt 28pt 12pt 12pt, clip,
    ]{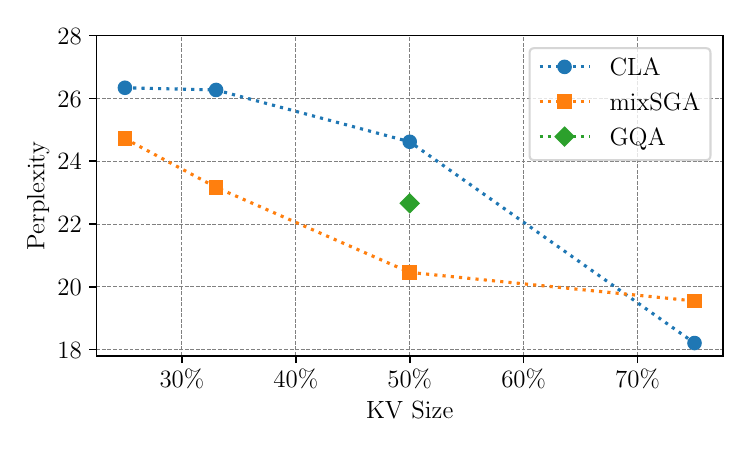}\cvspace{-0.5em}%
    \caption{%
        Comparing TinyLlama-1.1B continued pretraining
        with \Method{} and baselines (GQA and CLA)
        across varying KV size ratios.
        Lower perplexity indicates better language modeling performance.
    }\label{fig:ppl_vs_kv}\cvspace{-1em}
\end{figure}

As illustrated in \Cref{fig:ppl_vs_kv},
\Method{} consistently achieves superior performance,
manifesting in lower perplexity across most KV budgets
compared to the baselines.
Notably,
CLA experiences a pronounced increase
in perplexity as the KV budget decreases,
particularly below \( 50\% \),
where its performance deteriorates significantly.
This highlights the challenges
faced by static approaches in maintaining accuracy
under constrained KV budgets.
Conversely,
\Method{} exhibits enhanced robustness,
with lower perplexity levels across various budgets,
suggesting that its dynamic token routing mechanism
enables more effective resource allocation.
This adaptability underscores its capability
to deliver improved language modeling performance,
even under limited KV budgets.

%% file: tables/main.tex
\begin{table*}[ht]
    \centering\adjustbox{max width=\linewidth}{%
    \iftoggle{arxiv}{}{\renewcommand{\arraystretch}{0.8}}
    \begin{tabular}{ll|l||r@{}r|r@{}r|r@{}r|r@{}r|r|r}
        \toprule
        \multirow{2}{*}{Architecture} & \multirow{2}{*}{Method}
            & Expert
            & \multicolumn{2}{c|}{\textbf{Dolly}}
            & \multicolumn{2}{c|}{\textbf{Self-Instruct}}
            & \multicolumn{2}{c|}{\textbf{Vicuna}}
            & \multicolumn{2}{c|}{\textbf{SN}}
            & \multicolumn{1}{c|}{\textbf{UN}}
            & \multicolumn{1}{c}{\textbf{Avg\@.}} \\
        & & Ratios
            & R-L & \hspace{0.25em}DSv3
            & R-L & \hspace{0.25em}DSv3
            & R-L & \hspace{0.25em}DSv3
            & R-L & \hspace{0.25em}DSv3
            & R-L & R-L \\
        \midrule
        \multirow{3}{*}{OPT-125m}
        & GQA & \r010
            & 17.70 & 2.19 & 6.93 & 2.35 & 12.57 & 1.81 & 8.33 & 2.28 & 10.56 & 11.22 \\
        & \Method{} & \r112
            & 19.80 & 3.22 & 6.89 & 3.22 & 12.33 & 2.16 & 11.21 & 2.65 & 13.77 & \textbf{12.80} \\
        & \Method{} & \r316
            & 17.65 & 3.25 & 7.75 & 2.96 & 10.79 & 2.48 & 8.77 & 2.99 & 12.83 & 11.56 \\
        \midrule
        \multirow{3}{*}{OPT-355m}
        & GQA & \r010
            & 21.11 & 3.19 & 7.88 & 2.09 & 10.86 & 1.75 & 10.51 & 2.27 & 12.77 & 12.63 \\
        & \Method{} & \r112
            & 17.21 & 3.36 & 9.19 & 3.06 & 10.55 & 2.10 & 11.03 & 2.72 & 14.67 & 12.53 \\
        & \Method{} & \r316
            & 21.43 & 3.48 & 8.68 & 3.57 & 12.19 & 2.64 & 11.34 & 2.85 & 14.90 & \textbf{13.71} \\
        \midrule
        \multirow{3}{*}{Llama3.2-1B}
        & GQA & \r010
            & 20.09 & 3.45 & 7.90 & 3.17 & 13.21 & 2.41 & 12.43 & 2.74 & 14.50 & 13.63 \\
        & \Method{} & \r112
            & 18.87 & 4.02 & 9.01 & 3.68 & 10.93 & 2.97 & 14.09 & 3.33 & 17.70 & 14.12 \\
        & \Method{} & \r316
            & 20.11 & 4.05 & 10.03 & 3.65 & 14.41 & 2.99 & 15.52 & 3.24 & 20.42 & \textbf{16.10} \\
        \midrule
        \multirow{3}{*}{Llama3.2-3B}
        & GQA & \r010
            & 23.26 & 4.19 & 9.95 & 3.45 & 14.93 & 35.4 & 15.73 & 3.68 & 18.23 & 16.42 \\
        & \Method{} & \r112
            & 25.49 & 5.08 & 11.20 & 4.66 & 15.34 & 4.29 & 19.46 & 4.12 & 24.19 & 19.14 \\
        & \Method{} & \r316
            & 25.57 & 5.23 & 13.13 & 4.43 & 14.61 & 3.86 & 18.32 & 4.18 & 24.24 & \textbf{19.17} \\
        \midrule
        \multirow{3}{*}{Llama3.1-8B}
        & GQA & \r010
            & 27.40 & 4.85 & 11.60 & 4.60 & 15.36 & 3.43 & 21.72 & 4.22 & 23.75 & 19.97 \\
        & \Method{} & \r112
            & 26.50 & 6.40 & 17.22 & 6.01 & 15.06 & 4.90 & 32.52 & 6.43 & 33.91 & 25.04 \\
        & \Method{} & \r316
            & 28.47 & 6.97 & 17.30 & 5.93 & 19.19 & 4.88 & 35.81 & 6.68 & 34.62 & \textbf{27.08} \\
        \midrule
        \multirow{3}{*}{Gemma2-2B}
        & GQA & \r010
            & 25.68 & 5.64 & 10.43 & 3.73 & 16.53 & 4.25 & 20.00 & 4.27 & 23.68 & 19.26 \\
        & \Method{} & \r112
            & 24.79 & 6.18 & 16.08 & 5.37 & 12.70 & 5.26 & 26.01 & 5.55 & 27.39 & 21.39 \\
        & \Method{} & \r316
            & 26.15 & 6.25 & 17.36 & 5.62 & 14.47 & 5.40 & 26.82 & 5.98 & 28.71 & \textbf{22.70} \\
        \bottomrule
    \end{tabular}}\cvspace{-0.5em}
    \caption{%
        Supervised fine-tuning of a range of models
        on the Dolly-15k instruction-following dataset \cite{DatabricksBlog2023DollyV2}.
        Evaluation includes ROUGE-L (R-L)
        and DeepSeek-V3 feedback scores (DSv3)
        on 5 conversational datasets.
        \Method{}
        demonstrates consistent improvements over GQA baselines
        with the same KV budgets.
        The ``Avg\@. R-L'' column
        shows the average ROUGE-L scores
        across all datasets.
    }\label{tab:main}
    \cvspace{-1em}
\end{table*}

%% file: tables/tiny_llama_results.tex
\begin{table}[t]
    \centering
    \adjustbox{max width=\linewidth}{%
    \begin{tabular}{l||ccccc|c|c}
    \toprule
    \iftoggle{arxiv}{%
        \textbf{Method}
        & \textbf{\upar HellaSwag}
        & \textbf{\upar PIQA}
        & \textbf{\upar Winogrande}
        & \textbf{\upar ARC-E}
        & \textbf{\upar ARC-C}
        & \textbf{\upar Average}
        & \textbf{\downar Wikitext-2} \\
    }{%
        \textbf{Method}
        & \textbf{\upar HS}
        & \textbf{\upar PI}
        & \textbf{\upar WG}
        & \textbf{\upar AE}
        & \textbf{\upar AC}
        & \textbf{\upar Avg.}
        & \textbf{\downar WT} \\
    }
    \midrule
    GQA & 36.70 & \textbf{70.62} & 55.90 & 54.92 & 23.89 & 48.41 & 22.66 \\
    CLA & 35.90 & 68.82 & 55.40 & \textbf{55.47} & 23.81 & 47.88 & 24.62 \\
    \Method{} & \textbf{37.00} & 69.53 & \textbf{56.30} & 54.84 & \textbf{25.17} & \textbf{48.57} & \textbf{20.46} \\
    \bottomrule
    \end{tabular}}\cvspace{-0.5em}%
    \caption{%
        Continued pretraining
        on TinyLlama-1.1B with MiniPile.
        (\upar: higher is better, \downar: lower is better%
        \iftoggle{arxiv}{}{,
            HS: HellaSwag, PI: PIQA, WG: Winogrande,
            AE: ARC-E, AC: ARC-C, WT: Wikitext-2}.)
    }\label{tab:tiny_llama_results}
\end{table}

%% file: tables/mixsga_h2o.tex
\begin{table}[!h]
    \centering
    \adjustbox{max width=\linewidth}{%
    \begin{tabular}{@{}ll||cccccc|c@{}}
    \toprule
    \iftoggle{arxiv}{%
        \textbf{H2O KR} & \textbf{Method} & \textbf{\upar HellaSwag} & \textbf{\upar PIQA} & \textbf{\upar Winogrande} & \textbf{\upar ARC-E} & \textbf{\upar ARC-C} & \textbf{\upar Average} & \textbf{\downar Wikitext-2} \\
    }{%
        \textbf{KR} & \textbf{Method} & \textbf{\upar HS} & \textbf{\upar PI} & \textbf{\upar WG} & \textbf{\upar AE} & \textbf{\upar AC} & \textbf{\upar Avg.} & \textbf{\downar WT} \\
    }
    \midrule
    \multirow{2}{*}{80\%}
        & GQA & \textbf{36.8} & 69.58 & 55.04 & 53.41 & 23.81 & 47.73 & 22.70 \\
        & \Method{} & 36.5 & \textbf{70.02} & \textbf{55.33} & \textbf{53.91} & \textbf{25.77} & \textbf{48.31} & \textbf{20.53} \\
    \midrule
    \multirow{2}{*}{60\%}
        & GQA & 36.3 & 68.62 & \textbf{54.06} & 53.17 & 23.63 & 47.16 & 22.72 \\
        & \Method{} & \textbf{36.5} & \textbf{70.18} & 53.51 & \textbf{53.66} & \textbf{25.34} & \textbf{47.84} & \textbf{20.63} \\
    \midrule
    \multirow{2}{*}{40\%}
        & GQA & \textbf{36.1} & 67.94 & 53.27 & 52.19 & 24.40 & 46.78 & 22.80 \\
        & \Method{} & 35.8 & \textbf{69.10} & \textbf{54.14} & \textbf{52.27} & \textbf{24.92} & \textbf{47.25} & \textbf{20.98} \\
    \midrule
    \multirow{2}{*}{20\%}
        & GQA & 35.2 & 63.18 & 49.96 & 44.36 & 21.33 & 42.81 & 23.56 \\
        & \Method{} & \textbf{35.5} & \textbf{64.80} & \textbf{50.75} & \textbf{44.51} & \textbf{22.53} & \textbf{43.62} & \textbf{22.19} \\
    \bottomrule
    \end{tabular}}\cvspace{-0.5em}%
    \caption{%
        Integrating H2O with various KV keep ratios
        on Gemma2-2b.
        \Method{} consistently outperforms GQA
        across most tasks and H2O KV keep ratios (KR).
    }\label{tab:mixsga_h2o}
\end{table}

%% file: tables/effect_ratios.tex
\begin{table}[t]
    \centering
    \adjustbox{max width=\linewidth}{%
    \begin{tabular}{l||ccccc|c}
        \toprule
        \iftoggle{arxiv}{%
            \textbf{Ratios}
            & \textbf{Dolly} & \textbf{Self-Instruct} & \textbf{Vicuna}
            & \textbf{Super-Natural} & \textbf{Unnatural} & \textbf{Avg\@.}
        }{%
            \textbf{Ratios}
            & \textbf{DL} & \textbf{SI} & \textbf{VC}
            & \textbf{SN} & \textbf{UN} & \textbf{Avg.}
        } \\
        \midrule
        \r192 & 18.41 & 7.80 & 11.49 & 9.78 & 13.53 & 12.20 \\
        \r162 & \textbf{19.60} & 8.47 & \textbf{12.14} & 11.53 & 14.79 & 13.31 \\
        \r112 & 18.87 & \textbf{9.01} & 10.93 & \textbf{14.09} & \textbf{17.70} & \textbf{14.12} \\
        \bottomrule
    \end{tabular}}\cvspace{-0.5em}%
    \caption{%
        Effect of different expert group ratios
        under the same KV size budget (50\%) for Llama3.2-1B.
        Results are reported for ROUGE-L across multiple benchmarks.
        (DL: Dolly Evaluation,
        SI: Self-Instruct,
        VC: Vicuna,
        SN: Super-Natural Instructions,
        UN: Unnatural Instructions,
        Avg.: Average ROUGE-L across benchmarks)
    }\label{tab:ablation:ratios}
\end{table}

%% file: tables/effect_kv_cache.tex
\begin{table}[t]
    \centering
    \adjustbox{max width=\linewidth}{%
    \begin{tabular}{l||ccccc|c}
        \toprule
        \iftoggle{arxiv}{%
            \textbf{Ratios}
            & \textbf{Dolly} & \textbf{Self-Instruct} & \textbf{Vicuna}
            & \textbf{Super-Natural} & \textbf{Unnatural} & \textbf{Avg\@.}
        }{%
            \textbf{Ratios}
            & \textbf{DL} & \textbf{SI} & \textbf{VC}
            & \textbf{SN} & \textbf{UN} & \textbf{Avg.}
        } \\
        \midrule
        \r316
            & \textbf{20.11} & \textbf{10.03} & \textbf{14.41} & \textbf{15.52} & \textbf{20.42} & \textbf{16.10} \\
        \r340
            & 18.11 & 9.12 & 15.47 & 12.78 & 16.25 & 14.35 \\
        \midrule
        \r112
            & \textbf{18.87} & \textbf{9.01} & \textbf{10.93} & \textbf{14.09} & \textbf{17.70} & \textbf{14.12} \\
        \r120
            & 13.94 & 6.83 & 14.75 & 8.80 & 11.06 & 11.08 \\
        \bottomrule
    \end{tabular}}\cvspace{-0.5em}%
    \caption{%
        Effect of redistributing KV cache across tokens
        under fixed KV size (50\% of the original model) for Llama3.2-1B.
        Results are reported for ROUGE-L
        following the style in \Cref{tab:ablation:ratios}.
    }\label{tab:ablation:expert_count}\cvspace{-0.5em}
\end{table}

%% file: tables/auxiliary_loss.tex
\begin{table}[t]
    \centering
    \adjustbox{max width=\linewidth}{%
    \begin{tabular}{l||cccc|c}
        \toprule
        \iftoggle{arxiv}{%
            \textbf{Ratios}
            & \textbf{Dolly} & \textbf{Self-Instruct} & \textbf{Vicuna}
            & \textbf{Super-Natural} & \textbf{Avg\@.}
        }{%
            \textbf{Ratios}
            & \textbf{DL} & \textbf{SI} & \textbf{VC}
            & \textbf{SN} & \textbf{Avg.}
        } \\
        \midrule
        \Method{} & \textbf{26.15} & \textbf{17.36} & \textbf{14.47} & \textbf{26.82} & \textbf{21.20} \\
        Random router & 24.65 & 12.56 & 12.24 & 20.98 & 17.68 \\
        No auxiliary loss & 10.07 & 6.22 & 4.56 & 8.54 & 7.35 \\
        \bottomrule
    \end{tabular}}\cvspace{-0.5em}%
    \caption{%
    Ablation study
    on the effect of auxiliary loss
    and learned routing for Gemma2-2B
    with \r316 expert ratios under a 50\% KV budget.
    Results report ROUGE-L scores across benchmarks.
}\label{tab:ablation:routing}\cvspace{-0.5em}
\end{table}

%% file: conclusion.tex
\section{Conclusion}\label{sec:conclusion}

This paper introduced \Method{},
a framework that combines dynamic token-wise expert-choice routing
with attention grouping
to optimize KV representations.
By using weight-shared heterogeneous attention experts,
\Method{} adaptively allocates resources based on token importance.
Our experiments with Llama3, OPT, and Gemma2 model families
show that \Method{} outperforms baseline approaches
in computational efficiency and task performance,
with improved scalability in resource-constrained scenarios.
The routing mechanism of \Method{}
ensures consistency between prefill and decode phases.
Overall,
\Method{} offers a scalable and efficient solution
for dynamic KV optimization.

%% file: limitations.tex
\section{Limitations and Risks}

Our method assigns tokens to diverse experts
within each layer to improve efficiency.
However,
we only configure the same capacity ratios
across different layers,
which may ignore diverse token importance
across lower and deeper layers.
Therefore,
our method can be further improved
with a global importance metric,
automatically configuring capacity ratios
tailored for each layer.
This potentially offers
versatility and flexibility
for better resource utilization.
Moreover,
mixSGA's efficiency gains in computational and memory costs
enable resource-constrained researchers to leverage advanced LLMs,
fostering innovation in fields like NLP and healthcare
while supporting sustainable AI through reduced energy consumption.
This democratization of access
can accelerate scientific progress and broaden AI's societal benefits.
However,
these efficiencies could lower barriers for harmful uses,
such as generating misinformation or amplifying biases in routing decisions,
and may introduce vulnerabilities like slowdowns from adversarial inputs.
Future work may incorporate fairness-aware routing
and adversarial robustness to ensure ethical deployment,
aligning technological advances with responsible AI practices.

%% file: appendix.tex
\section{Experimental Setup}\label{app:setup}
Our experiments are conducted on
open-sourced datasets.
These datasets serve as artifacts
for research purposes,
which is aligned with
the goal of our experimental evaluation.
Specifically,
Databricks-Dolly-15k dataset uses CC BY-SA 3.0 license.
Wikitext-2 is available under the Creative Commons Attribution-ShareAlike License.
The remaining datasets in lm-eval-harness
are available under the MIT License.

\subsection{Supervised Fine-Tuning Tasks}\label{app:setup:sft}

For the supervised fine-tuning tasks,
we apply templates to
both the training and test datasets,
following the standard procedure
described in \cite{gu2024,kodistillm}.
All input text was standardized
to ensure consistency and fairness
across different models.
For the DeepSeek-V3 feedback evaluation \cite{deepseekai2024dsv3},
we use the template shown in \Cref{fig:feedback_template},
with a temperature coefficient
set to 0.7 to
balance the randomness and diversity
of the generated outputs.
We first construct the training data
from the Databricks-Dolly-15k dataset \cite{DatabricksBlog2023DollyV2},
wherein we randomly select 14,000 samples
for training and equally leave 500 samples
for validation and testing, respectively.

As part of our baselines,
we modify the original pretrained models by
integrating them into a more advanced GQA setup.
For models not originally including GQA results,
we apply the GQA mechanism.
In cases where
the models already have GQA results,
we replace them with a more compressed version of GQA,
which offers stronger compression levels.
This ensures that our baseline is consistently adapted
for a fair comparison with the new methods.

We performed full parameter fine-tuning for
the OPT model series (OPT-\{125m, 355m\}).
The batch size was set to 32,
and we used a cosine learning rate schedule.
The learning rate was initially set to \(5e^{-5}\) and
decayed according to the cosine decay scheduler.
The models were trained for 40 epochs
to ensure sufficient fine-tuning.

The routing weights were initialized
using He initialization \cite{he2015init}.
For the learning rate setup,
we initialized the learning rate at \(5e^{-5}\),
and the learning rate decay
followed the same cosine schedule.
We used a batch size of 32 for
both training and evaluation.
We employed gradient accumulation
to simulate a larger batch size
without exceeding memory constraints.


\subsection{Continued Pretraining Tasks}\label{app:setup:continued}

In the continued pretraining tasks,
we fine-tune the TinyLlama-1.1b model
using the MiniPile dataset \cite{kaddour2023minipile},
which contains 1.6 billion tokens.
The pretraining weights for
the TinyLlama model were originally trained
on a much larger dataset
containing 3 trillion tokens \cite{zhang2024tinyllama}.
To adapt it as our baseline,
we reduce the number of KV heads by half
and implement a deeper GQA configuration.
This modification of pretrained weights degrades
the original model's performance,
as halving the KV heads
impacts the model's integrity.
Therefore, we perform continued pretraining
on the 1.6 billion tokens of MiniPile
to recover the model's performance
and address this degradation.
Note that for all methods,
we use the same hyperparameter settings
in continued pretraining experiments,
as illustrated in \Cref{tab:train-hyperparameters-1.1B}
and \Cref{tab:model-hyperparameters-1.1B},
for fair comparison.


We train the TinyLlama-1.1B model
for one epoch on the MiniPile dataset
and then evaluate its performance
using lm-eval-harness\cite{eval-harness} framework
in a zero-shot setting
across several benchmarks,
including HellaSwag \cite{zellers2019hellaswag},
PIQA \cite{Bisk2020},
Winogrande \cite{sakaguchi2019winogrande},
ARC-Easy (ARC-E), ARC-Challenge (ARC-C) \cite{allenai_arc},
and perplexity on Wikitext-2 \cite{merity2016pointer}.
These benchmarks present comprehensive assessment of
the model's language modeling abilities
and task-specific performance.

\begin{figure}[h]
    \begin{framed}\tt\small
    Please act as an impartial judge and evaluate the quality of the response provided by an AI assistant to the user question displayed below.   Consider factors such as helpfulness, relevance, accuracy, depth, and creativity.   While evaluating, focus on the positive aspects, including clarity, usefulness, and effort, even if there are minor areas for improvement.  Please rate the response on a scale of 1 to 10 by following this format: 'Rating: [[x.xx]]', for example: 'Rating: [[5.00]].
    \end{framed}\caption{%
        System prompt template for DeepSeek-V3 feedback evaluation.
    }\label{fig:feedback_template}
\end{figure}
\input{tables/cft_setup.tex}

\input{tables/tiny_llama_model_setup.tex}

\section{Computational Resources}\label{app:resources}

The experiments were conducted on
two types of server equipped with
NVIDIA A100 and V100 GPUs,
configured by different model sizes and precision types.

For the Llama 8B model,
we used servers with 4 NVIDIA A100 GPUs (80GB)
and Intel Xeon Gold 6230R processors with 104 CPU cores.
We use bfloat16 (bf16) precision
to align with the precision applied for pretraining
and reduce memory burden.

For other Llama and Gemma model series,
our experiments were performed
on servers with 4 NVIDIA A100 GPUs (40GB)
and the same CPU and precision configurations.

The OPT models (OPT-\{125m, 355m\})
were trained on 4 NVIDIA V100 GPUs (32GB),
with Intel Xeon Gold 5118 processors with 48 CPU cores,
using float32 (fp32) precision
due to the V100's lack of
hardware support for bfloat16 format.




\section{Additional Experiments}\label{app:experiments}

\subsection{Compute and Memory Overheads}\label{app:experiments:overheads}

\Cref{tab:overhead} presents the real compute and memory overheads
of \Method{} compared to GQA.
It shows that \Method{} incurs a marginal increase in FLOPs and KV sizes,
with slightly higher parameter overheads due to routing weights.
To compare the real inference time
of \Method{} and GQA,
we measure decoding throughput (tokens per second)
under a 50\% KV budget,
using a batch size of 1
and generating 10,000 tokens
over five trials.
As shown in \Cref{tab:throughput},
\Method{} achieves throughput performance
of 57.28, 38.65, and 25.92 tokens/s
on Llama3.2-1b, Llama3.2-3b, and Llama3.1-8b, respectively,
compared to GQA's 59.75, 40.16, and 26.77 tokens/s.
This results in a modest 3-4\% overhead for \Method{},
reflecting its dynamic
routing complexity.
These results,
based on a na{\"\i}ve implementation,
suggest significant potential
for optimization to further reduce this gap.
\input{tables/overhead}
\input{tables/tps}


\subsection{Optimal Expert Ratio Analysis}\label{app:experiments:optimal_ratios}
To understand
why the \r316 expert ratio
consistently yields superior performance across models,
we analyze the allocation
of expert ratios under a fixed 50\% KV budget.
The ratios (\(x, y, z\)) for three experts
are constrained as follows:
\begin{equation}
    \left\{
    \begin{array}{ll}
    x + 0.5y + 0.25z = 0.5, & \quad \text{Budget,} \\
    x + y + z = 1, & \quad \text{Allocation,} \\
    0 \leq x, y, z \leq 1, & \quad \text{Bound.}
    \end{array}
    \right.
\end{equation}
We perform a grid search on Gemma2-2B,
varying \( x \) in \( 0.05 \) increments,
resulting in six feasible configurations
as shown in \Cref{tab:ablation:optimal_ratios}.
Notably,
\(x=0.3, y=0.1, z=0.6\),
(\ie{}, \( \r{x}{y}{z} = \r316 \))
achieves the highest
average ROUGE-L score (21.20)
\iftoggle{arxiv}{%
across Dolly, Self-Instruct, Vicuna,
and Super-Natural Instructions}{%
across Dolly (DL), Self-Instruct (SI), Vicuna (VC),
and Super-Natural Instructions (SN)},
outperforming other configurations.
This indicates that allocating 30\%
of tokens to the first expert (group size 1),
10\% to the second (group size 2),
and 60\% to the third (group size 3)
optimizes performance
by prioritizing significant tokens
for the first expert
while efficiently handling
less critical tokens with the third.
These results justify the adoption
of the default ratio for our experiments
in this study.
\input{tables/optimal_ratios.tex}

%% file: tables/cft_setup.tex
\begin{table}[ht]
    \centering
    \begin{tabular}{lc}
        \toprule
        \textbf{Model Size} & \textbf{1.1B} \\ \midrule
        Max LR & 2e-4 \\
        LR Scheduler & cosine \\
        Optimizer & AdamW \\
        $\beta_1$ & 0.9 \\
        $\beta_2$ & 0.95 \\
        Warmup Ratio & 0.015 \\
        Weight Decay & 0.1 \\
        Gradient Clipping & 1.0 \\
        Precision & Bfloat16 \\
        Batch Size (tokens) & 256K \\
        Epochs & 1 \\
        DataSet & MiniPile \\
        GPU & A100 \\ \bottomrule
    \end{tabular}
    \caption{%
        Training Hyperparameters for Continued Pre-training (TinyLlama-1.1B)
    }\label{tab:train-hyperparameters-1.1B}
\end{table}

%% file: tables/tiny_llama_model_setup.tex
\begin{table}[ht]
    \centering
    \begin{tabular}{lc}
        \toprule
        \textbf{Model Size} & \textbf{1.1B} \\ \midrule
        Hidden Size & 2048 \\
        Intermediate Size & 5632 \\
        Max Trained Length & 2048 \\
        \# Layers & 22 \\
        \# Attention Heads & 32 \\
        \# KV Heads & 4 \\
        RMS Norm eps & 1e-5 \\
        Vocab Size & 32000 \\ \bottomrule
    \end{tabular}
    \caption{%
        Model Hyperparameters for TinyLlama 1.1B
    }\label{tab:model-hyperparameters-1.1B}
\end{table}

%% file: tables/overhead.tex
\begin{table}[t]
\centering\adjustbox{max width=\linewidth}{%
\begin{tabular}{ll||rrr}
    \toprule
    \iftoggle{arxiv}{%
        \textbf{Metrics} & \textbf{Method}
        & \textbf{OPT-125m}
        & \textbf{Llama3.2-1b}
        & \textbf{Gemma2-2b}
    }{%
        \textbf{Metrics} & \textbf{Method}
        & \textbf{O-125m}
        & \textbf{L3.2-1b}
        & \textbf{G2-2b}
    }\\
    \midrule
    \multirow{2}{*}{\#Params}
        & GQA & 118m & 1.22b & 2.55b \\
        & \Method & 125m & 1.24b & 2.61b \\
    \midrule
    \multirow{2}{*}{\#FLOPs}
        & GQA & 94.8k & 113k & 237k \\
        & \Method & 94.9k & 113k & 237k  \\
    \midrule
    KV size per 
        & GQA & 36,864 & 32,768 & 106,496 \\
    token (bytes)
        & \Method & 36,867 & 32,772 & 106,502 \\
    \bottomrule
\end{tabular}}\cvspace{-0.5em}%
\caption{%
    Comparison of parameter counts,
    FLOPs,
    memory usage,
    and time for GQA and \Method{}
    under the same KV size budget (50\%).
    \iftoggle{arxiv}{%
    }{%
        (O-125m: OPT-125m,
        L3.1-8b: Llama3.1-8b,
        L3.2-1b: Llama3.2-1b,
        G2-2b: Gemma2-2b)}
}\label{tab:overhead}
\end{table}

%% file: tables/tps.tex
\begin{table}[t]
    \centering
    \begin{tabular}{l|ccc}
        \toprule
        \iftoggle{arxiv}{%
            \textbf{Method}
            & \textbf{Llama3.2-1b}
            & \textbf{Llama3.2-3b}
            & \textbf{Llama3.1-8b}
        }{%
            \textbf{Method}
            & \textbf{L3.2-1b} & \textbf{L3.2-3b} & \textbf{L3.1-8b}
        } \\
        \midrule
        GQA & 59.75 & 40.16 & 26.77 \\
        \Method{} & 57.28 & 38.65 & 25.92 \\
        \midrule
        $\Delta$ (\%) & 4.13 & 3.75 & 3.17 \\
        \bottomrule
    \end{tabular}
    \caption{%
    Decoding throughput (tokens/s) of GQA and \Method{} under a 50\% KV budget,
    with batch size 1, 10,000 tokens generated,
    and averaged over five trials. Higher is better.
    \iftoggle{arxiv}{%
    }{%
        (L3.2-1b: Llama3.2-1b,
        L3.2-3b: Llama3.2-3b,
        L3.1-8b: Llama3.1-8b)}
}\label{tab:throughput}
\end{table}

%% file: tables/optimal_ratios.tex
\begin{table}[h]
    \centering
    \adjustbox{max width=\linewidth}{%
    \begin{tabular}{l||cccc|c}
        \toprule
        \iftoggle{arxiv}{%
            \textbf{Ratios}
            & \textbf{Dolly} & \textbf{Self-Instruct} & \textbf{Vicuna}
            & \textbf{Super-Natural} & \textbf{Avg\@.}
        }{%
            \textbf{Ratios}
            & \textbf{DL} & \textbf{SI} & \textbf{VC}
            & \textbf{SN} & \textbf{Avg.}
        } \\
        \midrule
        \r1{17}2 & 20.61 & 10.18 & 6.42 & 14.54 & 12.94 \\
        \r172 & 23.49 & 10.70 & 14.10 & 17.72 & 16.50 \\
        \r3{11}6 & 25.36 & 12.60 & \textbf{16.90} & 18.52 & 18.35 \\
        \r122   & 25.80 & 13.45 & 15.54 & 19.04 & 18.46 \\
        \r112 & 24.79 & 16.08 & 12.70 & 26.01 & 19.90 \\
        \r316   & \textbf{26.15} & \textbf{17.36} & 14.47 & \textbf{26.82} & \textbf{21.20} \\
        \bottomrule
    \end{tabular}}
    \caption{%
    ROUGE-L scores (\upar) for Gemma2-2B
    with varying expert ratios
    under a 50\% KV budget
    \iftoggle{arxiv}{%
    across Dolly, Self-Instruct, Vicuna,
    and SuperNatural Instructions}{%
    across Dolly (DL), Self-Instruct (SI), Vicuna (VC),
    and SuperNatural Instructions (SN)}.
}\label{tab:ablation:optimal_ratios}
\end{table}